\pdfoutput=1

\documentclass[11pt,hyphens]{article}

\usepackage{EMNLP2023}
\usepackage{times}
\usepackage{enumitem}
\usepackage{multirow}
\usepackage{multicol}
\usepackage{float}
\usepackage{latexsym}

\usepackage{booktabs} 
\usepackage[normalem]{ulem}
\usepackage{xcolor} 
\usepackage{algorithm}
\usepackage[noend]{algpseudocode}
\usepackage{amsmath}
\usepackage{mathrsfs}
\usepackage{amssymb}
\usepackage{subfigure}
\usepackage{mathtools}
\usepackage[font=rm]{caption}
\DeclareCaptionType{copyrightbox}
\usepackage{shortvrb}
\usepackage{tabularx}
\usepackage{verbatim}
\usepackage{xspace}
\usepackage{listings}
\usepackage{tikz}
\usepackage{verbatim}
\usetikzlibrary{trees}
\usepackage{graphicx}
\usepackage{lipsum}

\makeatletter
\newcommand\blfootnote[1]{%
  \begingroup
  \renewcommand{\thefootnote}{}%
  \renewcommand{\@makefnmark}{}%
  \let\Hy@footnote@currentHref\@empty  
  \footnotetext[0]{#1}%
  \endgroup
}
\makeatother

\usepackage{fontawesome}
\usepackage[multiple]{footmisc}
\usepackage[all]{nowidow}
\usepackage{balance}
\usepackage{microtype}
\usepackage{etoolbox} 

\usepackage{inconsolata}

\title{Unsupervised Morphological Tree Tokenizer}

\author{Qingyang Zhu$^{1\ast}$,\ Xiang Hu$^{2\ast\dagger}$,\ Pengyu Ji$^{3}$,\ Wei Wu$^{2\dagger}$,\ Kewei Tu$^{3\dagger}$\vspace{5pt}\\
$^1$New York University,\ $^2$Ant Group,\ $^3$ShanghaiTech University\vspace{2pt}\\
 \tt{\normalsize{qz2457@nyu.edu}}\\
 \tt{\normalsize{\{aaron.hx, congyue.ww\}@antgroup.com}}\\
\tt{\normalsize{\{jipy2023, tukw\}@shanghaitech.edu.cn}} \\
}

\begin{document}
\maketitle
\blfootnote{$^\ast$ Equal contribution.}
\blfootnote{$^\dagger$ Corresponding authors.}
\begin{abstract}
As a cornerstone in language modeling, tokenization involves segmenting text inputs into pre-defined atomic units. Conventional statistical tokenizers often disrupt constituent boundaries within words, thereby corrupting semantic information. 
To address this drawback, we introduce morphological structure guidance to tokenization and propose a deep model to induce character-level structures of words.
Specifically, the deep model jointly encodes internal structures and representations of words with a mechanism named \textit{MorphOverriding} to ensure the indecomposability of morphemes. By training the model with self-supervised objectives, our method is capable of inducing character-level structures that align with morphological rules without annotated training data.
Based on the induced structures, our algorithm tokenizes words through vocabulary matching in a top-down manner. 
Empirical results indicate that the proposed method effectively retains complete morphemes and outperforms widely adopted methods such as BPE and WordPiece on both morphological segmentation tasks and language modeling tasks. Code is available at \url{https://github.com/martianmartina/TreeTokenizer}\footnote{Part of the work was done during Qingyang's internship at Ant Group.}.

\end{abstract} 

\section{Introduction}
Tokenization, the initial step of language modeling, segments natural language into manageable units. While this process is crucial for representing natural language, research on new tokenization methods has remained limited, particularly in contrast to the rapid advancements in language model architectures and learning approaches.
Currently, the de-facto tokenizers are BPE~\cite{sennrich-etal-2016-neural} and WordPiece~\cite{Schuster2012JapaneseAK}, which 
have been widely adopted by state-of-the-art language models such as GPT~\cite{radford2019language} and BERT~\cite{devlin-etal-2019-bert}.  However, numerous studies have challenged these methods~\cite{bostrom-durrett-2020-byte,Church_2020,hofmann-etal-2021-superbizarre,minixhofer-etal-2023-compoundpiece}, arguing that they cannot adequately capture linguistic information. They often disrupt constituent boundaries within words, leading to unnatural and fragmented token representations.
\begin{figure}[tb!]
    \centering
    \includegraphics[width=0.5\textwidth]{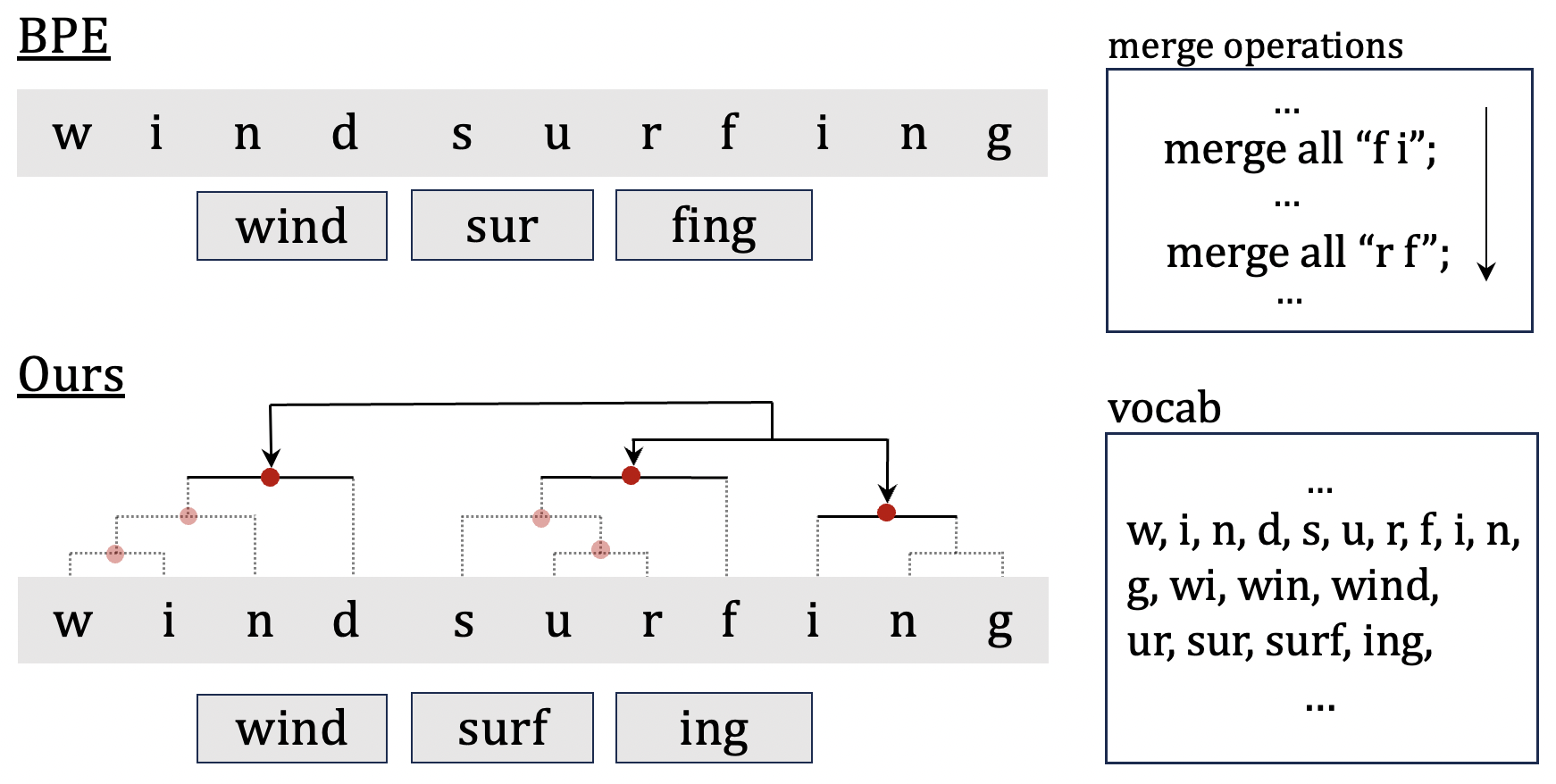}
    \vspace{-10pt}
    \caption{
    BPE (top) tokenizes a word through a bottom-up greedy merging approach given pre-learned merge operations, 
    while ours (bottom) tokenizes a word via a top-down vocabulary matching while traversing a global parse tree.}
    \label{fig:intuition}
    \vspace{-10pt}
\end{figure}
Figure~\ref{fig:intuition}(top) demonstrates an example where BPE fails to identify the appropriate boundaries in a word. 

According to linguistic theories, both words and sentences are believed to have internal structures~\cite{selkirk1982syntax, Marvin2002TopicsIT,DBLP:conf/naacl/CotterellS15}. While sentence-level grammar induction methods based on deep neural networks are highly effective, whether these methods can be applied equally well to words remains under-explored. In this work, we systematically evaluate neural grammar induction methods at the word level, propose a hypothesis explaining their suboptimal performance, and empirically validate this hypothesis. Building on these insights, we introduce the first effective unsupervised neural grammar induction model at the word level and present a more morphologically aligned tokenizer that leverages our model as shown in Figure ~\ref{fig:intuition}(bottom).

Our approach draws inspiration from syntactic composition models~\cite{DBLP:journals/corr/MaillardCY17}, where a sentence is encoded as a weighted sum over all composed root representations of its underlying binary parse trees via dynamic programming.  
Instead of composing a sentence from words, we apply composition models on characters in a word to induce its morphological parse tree.
To train the composition model, we propose two self-supervised objectives akin to next token prediction and span prediction that effectively leverage both contextual information at the sentence level and semantic information at the subword level.
Thus the model can learn to assign higher probabilities to morphological constituents of a word and induce the underlying morphological parse tree. 
 
However, character sequences present a unique challenge to composition models because morphemes, the smallest meaning-bearing units in a language~\cite{Jurafsky:2009:SLP:1214993}, are indecomposable.
While we can represent a constituent by composing its sub-constituents in most cases, we cannot represent a subword by composing its components if the subword is a morpheme.
For example, the meaning of \texttt{windsurf} can be decomposed to \texttt{wind+surf}, but \texttt{wind} is a morpheme whose meaning is not a function of its components. 
To address the challenge, we propose a mechanism named \textit{MorphOverriding}.
During the bottom-up composition process in our model, upon identifying a subword that matches an entry in a heuristically constructed morpheme vocabulary, we compute the subword representation from both its components and the corresponding morpheme embedding, i.e., the model may learn to mix or override the composition with the morpheme embedding.
Our experiments show that such a mechanism is critical in morphological structure induction. 

Building upon the resolution of morphological structure induction, we introduce a novel tokenization algorithm named \textit{TreeTok}, which includes both vocabulary construction and word segmentation. 
During vocabulary construction, TreeTok first utilizes a tree-based BPE variant to build an initial vocabulary and then applies a tree-based Unigram variant to prune the initial vocabulary to a specified size. 
Because TreeTok operates in a top-down manner, it does not need to retain all intermediate tokens produced by merge operations in the vocabulary as BPE does. By this means, we can build a more compact vocabulary by pruning less important subwords. During word segmentation, we employ a lightweight parser with compact parameters distilled from the composition model to parse a word into a character-level binary tree and then apply top-down vocabulary matching to enhance the tokenizer's alignment to morphological structure, as illustrated in Figure~\ref{fig:intuition}. 

In our experiments, which mainly focus on English, we train TreeTok and baselines on the Wikitext-103 corpus~\cite{mcclosky-etal-2006-effective} and assess their performance on morphological segmentation tasks and language modeling tasks. Evaluation results indicate that TreeTok consistently outperforms BPE and WordPiece across all the tasks. 

In conclusion, our contributions are three-fold:
\begin{itemize}[leftmargin=*,noitemsep,nolistsep]
\item We conduct empirical study on character-level neural parsing, identifying its limitations and proposing a novel explanation—lack of MorphOverriding—to account for its suboptimal performance.
\item Building on the MorphOverriding hypothesis, we introduce the first effective unsupervised neural model for character-level structure induction, addressing a critical gap in the field.
\item We show that our character-level structure induction method can be integrated into mainstream tokenizers to significantly enhance their performance on morphological tasks.
\end{itemize}
\section{Related Work}
\paragraph{Subword Tokenizers.}
Subword tokenization, with typical methods such as BPE~\cite{sennrich-etal-2016-neural} and WordPiece~\cite{Schuster2012JapaneseAK}, has become customary in most NLP fields. 
BPE builds its vocabulary by repeatedly merging the most frequent subword unit pairs, whereas WordPiece selects pairs using the highest mutual information. 
During tokenization, BPE applies learned merge operations in the same order to new text initialized with characters while WordPiece iteratively finds the longest match in the vocabulary.
Unigram~\cite{kudo-2018-subword}, another popular tokenizer, builds its vocabulary in the opposite direction: it starts with a large set of potential subwords and prunes them based on delta entropy in a unigram language model.
Our tokenizer aims to build upon the advantages of these effective statistical tokenizers and augment them with unsupervised induced tree structures.
\vspace{-5pt}
\paragraph{Unsupervised Morphological Segmentation.}
In the line of work on unsupervised morphological segmentation, the most well-known model is Morfessor~\cite{creutz-lagus-2002-unsupervised}, along with its multiple variants~\cite{6096b95d90a24172954c7dd770d83ab4, gronroos-etal-2014-morfessor, gronroos-etal-2020-morfessor}. In Morfessor, an online search algorithm is utilized to apply a hierarchical word splitting strategy with a Minimum Description Length (MDL)~\cite{Rissanen1989StochasticCI} cost function. However, its lack of explicit control over vocabulary size makes it unsuitable for use as a tokenizer. In addition, although some studies~\cite{DBLP:conf/amta/AtamanF18,DBLP:conf/emnlp/HouKVY23} find morphologically motivated segmentation can improve data-driven tokenizers, most other studies~\cite{DBLP:conf/tsd/MachacekVB18,DBLP:conf/cicling/DomingoGHCH19,DBLP:conf/eacl/SalevaL21} find no reliable improvement of such methods over BPE. According to \citet{galle-2019-investigating}, the effectiveness of BPE lies in its superior compression capability. A more detailed discussion can be found in \citet{mielke2021words}.
Some other studies try to model morphological structures using Bayesian PCFGs~\cite{johnson-etal-2007-bayesian} or a non-parametric Bayesian generalization of PCFGs~\cite{NIPS2006_62f91ce9}. 
However, they are pure statistical models and do not utilize modern neural methodologies.
Our method differs from previous unsupervised morphological methods in our character-based structures, thereby possessing the superior compression capability of BPE.
Meanwhile, our method leverages modern neural methodologies to better utilize contextual and intra-word semantic information.

\paragraph{Composition Model.} 
In this work, we utilize a composition model to induce morphological structures.
Composition models jointly learn representations and structures of a symbol sequence by transforming text encoding into a combinatorial optimization problem.
\newcite{DBLP:journals/corr/MaillardCY17} proposes a CKY-like~\cite{10.5555/1097042,kasami1966efficient,younger1967recognition} encoder, in which each constituent is represented as a weighted average of the set of composed representations computed from different splits of the constituent.
\newcite{drozdov-etal-2019-unsupervised-latent} proposes a deep inside-outside encoder~\cite{Baker1979TrainableGF,LARI199035}, enabling the encoder to learn underlying structures via an auto-encoding objective. 
Recently, a series of studies~\cite{hu2024generative,hu2024augmenting} have been conducted to reduce the deep inside-outside encoder complexity from cubic to linear, on which our work is based.
\section{Methodology}
To tokenize a word $\mathbf{x}=\{x_1, x_2, ..., x_n\}$ where $x_i$ is the $i$-{th} character, we aim to parse it into a binary tree and then tokenize it via top-down vocabulary matching. The parser is a deep composition model capable of jointly modeling the internal structures and representations of words. It is trained on sequences of words sampled from the corpus and optimized with self-supervised objectives that capture both intra-word compositionality and inter-word contextual dependencies.
In the following sections, we sequentially introduce the composition model, training objectives, and the tree-based tokenization algorithm.
\subsection{Composition Model for Word} 
\begin{figure}[tb!]
    \centering
    \includegraphics[width=0.50\textwidth]{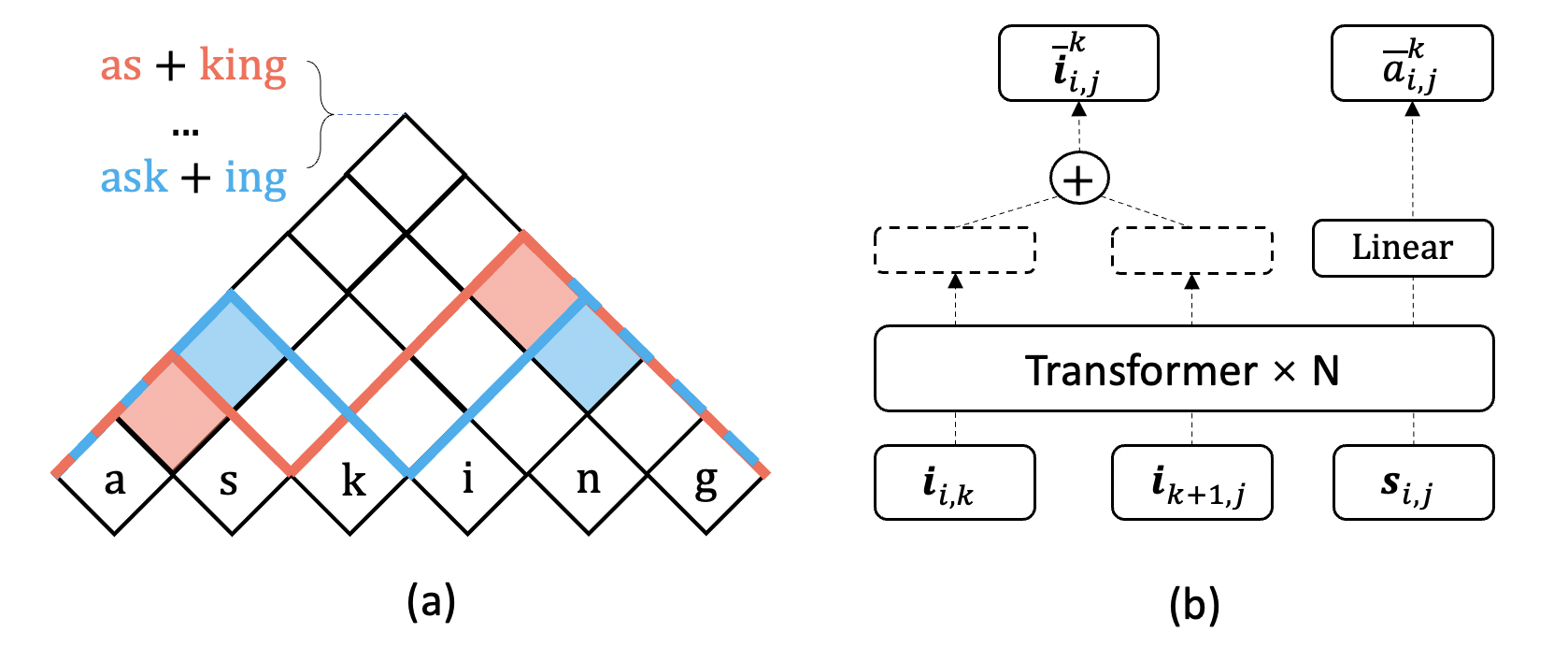}
    \vspace{-15pt}
    \caption{(a)
    The composition representation of \texttt{asking} ($\mathbf{i}_{1,6}$) is a weighted sum over all subword pairs such as \texttt{ask}+\texttt{ing} ($\bar{\mathbf{i}}_{1,6}^3$) and \texttt{as} + \texttt{king} ($\bar{\mathbf{i}}_{1,6}^2$).
    (b) The composition function. Take \texttt{ask} ($\mathbf{i}_{1,3}$) as an example. $\mathbf{s}_{1,3}$ is $\mathbf{E}_{\mathbb{V}[\texttt{ask}]}$ if \texttt{ask} $\in \mathbb{V}$. Thus the representation of \texttt{ask} depends not only on its components but also on $\mathbf{E}_{\mathbb{V}[\texttt{ask}]}$. However, if \texttt{asking} $\notin \mathbb{V}$, then $\mathbf{s}_{1,6}$ is $\mathbb{E}_{empty}$ and the representation of \texttt{asking} ($\mathbf{i}_{1,6}$) only depends on the composition representation of its components.}
    \label{fig:composition_func}
    \vspace{-10pt}
\end{figure}

For a given word $\mathbf{x}$, we denote $\mathbf{i}_{i,j}$ as the representation of subword $\mathbf{x}_{i:j}=\{x_i,...,x_j\}$. 
The inside pass~\cite{drozdov-etal-2019-unsupervised-latent} of a composition model computes a composition vector $\bar{\mathbf{i}}_{i,j}^k$ and a compatibility score $\bar{a}_{i,j}^k$ for each pair of sub-constituents $(i,k)$ and $(k+1,j)$. The compatibility score indicates how likely these two sub-constituents are to be merged. The constituent representation $\mathbf{i}_{i,j}$ is computed as a weighted average over composition vectors of all possible pairs of sub-constituents as follows:
\begin{equation}
\scalebox{0.9}{$
\begin{aligned}
&\bar{a}_{i,j}^k, \bar{\mathbf{i}}_{i,j}^k = f_{\alpha}(\mathbf{i}_{i,k}, \mathbf{i}_{k+1, j})\,,\\
&\,\hat{w}_{i,j}^k = \frac{\exp(\bar{a}_{i,j}^k)}{\sum_{k'=i}^{j-1}\exp(\bar{a}_{i,j}^{k'})}\,,\mathbf{i}_{i,j} = \sum_{k=i}^{j-1}\hat{w}_{i,j}^k\bar{\mathbf{i}}_{i,j}^k\,.
\end{aligned}
$}
\label{eq:neural_inside}
\end{equation}
The inside pass starts with characters by initializing $\mathbf{i}_{i,i}$ with character embeddings and recursively computes constituent representations bottom up following Equation~\ref{eq:neural_inside}. Representation $\mathbf{i}_{1,n}$ of the whole word $\mathbf{x}$ is regarded as the word embedding $\Call{emb}{\mathbf{x}}$. $f_\alpha$ is the composition function implemented with a multi-layered Transformer. 
An example of the bottom-up composition process is depicted in Figure~\ref{fig:composition_func}(a).
In this work, we employ a pruned version of deep inside encoder~\cite{hu2024augmenting} as our backbone, which is easy to scale up, thanks to the logarithmic parallel time complexity and the linear space complexity.

The limitation of this approach is that the representation of any subword is always composed of its component pairs, which is incompatible with the linguistic constraint that morphemes are the smallest meaning-bearing units and should not be decomposed further. Hence, we introduce MorphOverriding to enable a subword representation to disentangle from its component pairs when the subword is a morpheme.
Specifically, we construct a morpheme vocabulary $\mathbb{V}$ heuristically using a statistical method (BPE in this work), in which each entry is associated with a learnable vector in a morpheme embedding table $\mathbf{E}$. When $\mathbf{x}_{i:j}$ hits the vocabulary $\mathbb{V}$, we insert its morpheme embedding $\mathbf{s}_{i,j}$ into the computation of $\mathbf{i}_{i,j}$, making it possible to mix or override the composition vector with the morpheme embedding.
Thus, the composition vector and the compatibility scores can then be reformulated as:
\begin{equation*}
\scalebox{0.9}{$
\begin{aligned}
&\bar{a}_{i,j}^k, \bar{\mathbf{i}}_{i,j}^k = f_{\alpha}(\mathbf{i}_{i,k}, \mathbf{i}_{k+1, j}, \mathbf{s}_{i,j})\,,\\
&\mathbf{s}_{i,j}=\begin{cases}
\mathbf{E}_{\mathbb{V}[{\mathbf{x}_{i:j}}]}  & \text{ if } \mathbf{x}_{i:j} \in \mathbb{V} \\
\mathbf{E}_{empty}  & \text{ if } \mathbf{x}_{i:j} \notin \mathbb{V}
\end{cases}\,,\\
\end{aligned}
$}
\end{equation*}

Figure~\ref{fig:composition_func}(b) illustrates the composition function equipped with MorphOverriding.
Our experiments demonstrate that this mechanism is crucial for character-level structure induction.

\paragraph{Tree induction.}
For a given span $(i,j)$, the best split-point is $k$ with the highest compatibility score $\bar{a}_{i,j}^k$. Thus, to derive a parse tree, we can recursively select the best split-points top-down starting from the root span $(1,n)$. As the pruned inside-outside encoder produces a lightweight parser~\cite{hu-etal-2022-fast} with a compact parameter set as a byproduct, we use it for efficient inference during tokenization.

\subsection{Training Objectives}
The composition model is trained on \emph{contiguous sequences of words} sampled from the corpus rather than isolated words.
The overall loss for training the composition model is the summation of an intra-word auto-encoding loss $\mathcal{L}_{ae}$ and an inter-word auto-regression loss $\mathcal{L}_{ar}$.
The auto-encoding loss is based on predicting each character or morpheme from the rest of a word, leveraging intra-word structure.
The auto-regression loss is based on predicting next word in the sequence that leverages contextual cues to disambiguate different underlying structures of a word.
Under these objectives, the composition model learns to assign proper scores to each split point of a subword, benefiting from both intra-word compositionality and inter-word context.

\paragraph{Auto-encoding Loss (intra-word).}
Auto-encoding is a common practice of training a composition model. For our character-level composition model, we try to predict each character $x_i$ based on its neighboring context representations $\mathbf{i}_{1,i-1}$ and $\mathbf{i}_{i+1,n}$~\cite{hu-etal-2021-r2d2}. However, the auto-encoding objective turns out to be empirically ineffective when training our model probably because unlike word-level auto-encoding that requires selecting from tens of thousands of words in a vocabulary, here we only need to select from tens of characters, which is much less challenging.

To enhance learning efficacy, we propose predicting both individual characters and morphemes in the vocabulary $\mathbb{V}$. For instance, given the word \texttt{windsurf}, we mask out \texttt{wind} and let the model uncover the masked morpheme based on the visible part \texttt{\_surf}.
Analogous to the inside pass, the outside pass computes each outside representation $\mathbf{o}_{i,j}$ in a top-down manner based on context information outside span $(i,j)$, whose details are described in Appendix~\ref{appdx:neural_outside}.
we use $\mathbf{o}_{i,j}$ to predict each subword $\mathbf{x}_{i:j}$ that belongs to $\mathbb{V}$:
\begin{equation*}
\scalebox{0.9}{$
\begin{aligned}
&\mathcal{L}_{ae} = -\frac{1}{\mathcal{N}}\sum_{\mathbf{x}_{i:j}\in\mathbb{V}}^{}\log\frac{\exp(\mathbf{o}^T_{i,j}\mathbf{E}_{\mathbb{V}[\mathbf{x}_{i:j}]})}{\sum_{k=1}^{\left | \mathbb{V} \right | }{\exp(\mathbf{o}^T_{i,j}\mathbf{E}_{k})}}\,,
\end{aligned}
$}
\end{equation*}
where $\mathcal{N}$ is the total number of subwords belonging to the vocabulary.~\footnote{Subword spans in the vocabulary may overlap (e.g., “asking” contains both “ask” and “king”), resulting in multiple competing candidates for prediction. Among the overlapping candidates, intuitively, it should be more reasonable to predict a constituent than other subword candidates from the context. Therefore, we assign a constituency weight to each subword in the objective, as detailed in Appendix~\ref{appdx:span_weights}.
}
\paragraph{Auto-regression Loss (inter-word).}
Given a sentence $\mathbf{S}=\{\mathbf{x}^1, ...,\mathbf{x}^m\}$, whose word embedding is computed by the composition model,
we feed the composed word embeddings into a causal language model and let it pick the correct next word from candidates built via in-batch sampling for each step. 
Let $\mathbf{h}_t$ denote the $t$-{th} hidden states of the causal language model and $\mathcal{W}$ denote a deduplicated vocabulary built on all input words in the same batch, we have the auto-regression loss defined as:
\begin{equation*}
\scalebox{0.9}{$
\begin{aligned}
\mathcal{L}_{ar}=-\frac{1}{m}\sum_{t=1}^{m-1}\log \frac{\exp(\mathbf{h}_t \Call{emb}{\mathbf{x}^{t+1})}}{\sum_{\mathbf{x} \in \mathcal{W}}^{}\exp(\mathbf{h}_t\Call{emb}{\mathbf{x}})}\,.
\end{aligned}
$}
\end{equation*}

\subsection{Tokenization}
The proposed tree-based tokenization algorithm, TreeTok, includes segmentation and vocabulary construction procedures. As the latter depends on the former, we first discuss the segmentation procedure, followed by the vocabulary construction.
\paragraph{Segmentation Procedure.}
Given a constructed vocabulary, whose details are described later, we parse each word into a morphological tree and segment it via a \textit{top-down matching} approach, as illustrated in Figure~\ref{fig:intuition}(bottom).
Specifically, during the top-down traversal of a parse tree, we retain a subword and backtrack if the subword matches an entry in the vocabulary. 
Note that unsupervised structural learning is often imperfect, causing erroneous tokenization. For instance, an incorrect parse tree \texttt{[[[book]e]d]} may yield tokens \texttt{book e d} where \texttt{e d} should be merged. 
To address this issue, we propose a post-processing step to deal with mergeable pairs of segmented tokens. 
Specifically, we define the empirical probability of token $t$ as $\frac{\Call{count}{t}}{T}$, where $\Call{count}{t}$ is the frequency of $t$ in the entire corpus and $T=\sum_{t \in \mathbb{V}}^{}\Call{count}{t}$. Therefore, the probability of a certain merge is the production of the probabilities of all tokens.
We find the optimal merge by searching for the one with maximum probability among all potential merges via dynamic programming.
Detailed pseudo-code can be found in Appendix~\ref{appdx:tokenize}.

\paragraph{Vocabulary Construction.}
One drawback of BPE and WordPiece is that they have to keep all intermediate ``junk" tokens produced during the iterations of merge operations, which results in limited vocabulary space occupied by these meaningless tokens. For instance, if the corpus contains many occurrences of \texttt{low} and \texttt{lower}, the meaningless token \texttt{lo} will be added to the vocabulary before \texttt{low} and will not be removed later.
However, with the top-down matching framework, we don't need bottom-up merge operations to restore tokens, allowing us to prune unnecessary tokens and create a more compact vocabulary.
To build a compact vocabulary, we propose a vocabulary construction algorithm in which we employ a tree-based BPE-like algorithm to build a heuristic vocabulary and a tree-based Unigram algorithm to prune unnecessary subword units. Specifically, we initialize the token vocabulary with the character vocabulary and repeat the following steps to build a heuristic vocabulary given character-level tree structures of words:
\begin{enumerate}[leftmargin=*,noitemsep,nolistsep]
\item Count adjacent token pairs that share the same parent in the tree structure, e.g., given \texttt{[[b[o o]]k]}, only the pair (\texttt{o}, \texttt{o}) is counted.
\item Merge adjacent symbol pairs whose counts exceed a given threshold, e.g., \texttt{[[b[o o]]k]} $\to$ \texttt{[[b oo]k]}.
\item Repeat 1-2 until there are no new symbol pairs.
\end{enumerate}
In the pruning procedure, we start from the heuristic symbol vocabulary and prune it as follows:
\begin{enumerate}[leftmargin=*,noitemsep,nolistsep]
\item Tokenize the corpus via the top-down matching according to the current vocabulary. The total entropy of the whole corpus is defined as $\mathcal{H}_{\mathbb{V}}=-\sum_{t \in \mathbb{V}}^{}\frac{\Call{count}{t}}{T}\log\frac{\Call{count}{t}}{T}$ where $T=\sum_{t \in \mathbb{V}}^{}\Call{count}{t}$.
\item For each token $s$, calculate the entropy gain after removing that word from the vocabulary denoted as $\Delta\mathcal{H}_s=\mathcal{H}_{\mathbb{V}/\{s\}}-\mathcal{H}_{\mathbb{V}}$. Intuitively, the higher $\Delta\mathcal{H}_s$ is, the more important $s$ is.
\item Sort delta entropy of tokens and remove the lowest $k\%$ from $\mathbb{V}$. Repeat step 1-2 until $\left | \mathbb{V}  \right |$ reaches the target vocabulary size.
\end{enumerate}
In practice, we design a tree-based Viterbi algorithm~\cite{DBLP:journals/tit/Viterbi67} to implement the pruning procedure efficiently. The pseudo-code is presented in Appendix~\ref{appdx:vocab}.
\section{Experiments}
We focus on English in most of the experiments, but we also evaluated the composition model on Chinese in \ref{sec:tree_quality} and evaluated the tokenizer on German, which is a morphologically richer language than English, in a machine translation experiment in \ref{par:mt}. We evaluate the performance of TreeTok against the de-facto tokenizers such as BPE, WordPiece, and Unigram as primary baselines. 
\paragraph{Training setups.}
For a fair comparison, we train all tokenizers from scratch on the lowercase version of the WikiText-103 corpus~\cite{merity2017pointer} without any word boundary marker and set the same vocabulary size of 30,000. For BPE, WordPiece, and Unigram, we use the implementation and default training paradigm provided by the HuggingFace library\footnote{\url{https://github.com/huggingface/tokenizers}}. 
Regarding the composition model, we train it with a context window of up to 512 characters. 
We use GPT2 implemented from HuggingFace\footnote{\url{https://github.com/huggingface/transformers}} as our causal language model when computing the auto-regression loss. We present detailed configurations of our model and training setup in Appendix~\ref{appdx:model_config}. 

\paragraph{Evaluation datasets.}
We compare our tokenizer with other tokenizers for morphological alignment (detailed in \ref{sec:tok_quality}) using two datasets with gold-standard morphological segmentation. One is from the Morpho Challenge 2010 Workshop~\cite{kurimo-etal-2010-morpho} (Morpho), which contains 1,000 word forms with their segmentations corresponding to the surface forms of morpheme labels. The dataset contains instances of all kinds of morphological transformations, including inflection, derivation, and compounding. The other dataset is from \citet{minixhofer-etal-2023-compoundpiece} (Compound), which contains 759 compound words specifically designed to test the models' capabilities in decompounding. We also use these morphological segmentation datasets to evaluate the induced morphological parse trees (detailed in \ref{sec:tree_quality}).

In addition, we evaluate the tokenizers using statistical metrics that have been shown to strongly correlate with the performance on downstream tasks (see Table~\ref{tbl:renyi}). These metrics are calculated on the validation set of WikiText-103. 

\subsection{Tokenization Quality}\label{sec:tok_quality}
\paragraph{Metrics.}
We measure the performance of morphological segmentation via accuracy, i.e., the ratio of examples that are correctly segmented. We also consider a few statistical metrics that can directly assess the quality of tokenization, including Rényi Efficiency~\cite{zouhar-etal-2023-tokenization}, average sentence-level perplexity, and average number of tokens per sentence.
Rényi Efficiency is introduced by \citet{zouhar-etal-2023-tokenization} as a principled intrinsic measure of
tokenization quality and is claimed to yield a Pearson correlation of 0.78 with BLEU~\cite{papineni-etal-2002-bleu} on machine translation. Sentence-level perplexity is defined as $-\log\ p(\mathbf{s}) = -\sum_{i=1}^n \log\ p(s_i|s_{<i})$, where $\mathbf{s}=\{s_1, s_2, ..., s_n\}$ is a sentence with $s_i$ being the $i$-{th} token.
Since different tokenizers generate distinct segmentations leading to different numbers of tokens of the same word, sentence-level perplexity provides fairer evaluation compared with the default token-level perplexity $-\frac{1}{n} \log~p(\mathbf{s})$.

\begin{table}[tb!]
\begin{center}
\setlength{\tabcolsep}{2pt}
\resizebox{0.5\textwidth}{!}{
\begin{tabular}{l|c|c|c}
\hline \hline
                       & Morpho (Acc.) $\uparrow$ & Compound (Acc.) $\uparrow$ & $|\mathbb{V}|$\\ 
                       & EN.                     & EN.                      &                \\ \hline
BPE                    & 19.50                   & 62.98                    & 30,000         \\
WordPiece              & 26.20                   & 62.19                    & 30,000         \\
Unigram                & 27.10                   & 53.10                    & 30,000         \\
TreeTok                & \textbf{37.9}           & \textbf{68.07}           & 30,000         \\ 
\hline \hline
\end{tabular}
}
\end{center}
\caption{Results on two morphological segmentation datasets. This table can be seen as a comparison between tree-enhanced BPE (TreeTok) and vanilla BPE/WordPiece/Unigram.}
\vspace{-10pt}
\label{tbl:seg_acc}
\end{table}

According to Table~\ref{tbl:seg_acc}, TreeTok significantly surpasses BPE, WordPiece, and Unigram on the two morphological segmentation datasets.
The results demonstrate the efficacy of TreeTok in aligning with morphology. 

\paragraph{Rényi efficiency \& Perplexity.} 
\begin{table}[tbh!]
\begin{center}
\setlength{\tabcolsep}{2pt}
\resizebox{0.40\textwidth}{!}{
\begin{tabular}{l|cc|c|c}
\hline \hline
                       & Rényi$\uparrow$ & PPL$\downarrow$      & BLEU$\uparrow$ & avg. \#tokens       \\ \hline
BPE                    & 44.66 & 107.76  & 26.55 & 26.58         \\
WordPiece              & 44.54 & 110.97  &- & 26.60           \\
Unigram                & 45.07 & 106.91  & -& 31.68            \\
TreeTok                & 44.82 & 107.26  & 26.68 & 25.99 \\ \hline \hline
\end{tabular}
}
\end{center}
\vspace{-5pt}
\caption{Results for different tokenization models on WikiText103 with 30,000 vocabulary size.
}
\vspace{-10pt}
\label{tbl:renyi}
\end{table}

Table~\ref{tbl:renyi} reports the evaluation results in terms of Rényi efficiency and perplexity (PPL). TreeTok outperforms BPE and WordPiece on both Rényi and PPL. 
The improvements illustrate the benefits of TreeTok's structural constraints and more compact vocabulary. The tree structure constraints enable the segmentation of words into more morphology-aligned tokens, while the compact vocabulary allows for the inclusion of meaningful morphemes by removing intermediate tokens in the pruning process during vocabulary construction, under a top-down matching framework.
Unigram performs slightly better than TreeTok, but produces 22\% more tokens on average. 
A possible explanation for the better performance of Unigram is that Unigram tends to produce inflectional suffixes such as ``\texttt{-ing}" and ``\texttt{-ly}", while other methods tend to retain entire words. This difference makes it easier for Unigram to share the same stems and affixes between different word forms, thus achieving better parameter sharing.
However, under the Transformer architecture, an additional 22\% number of tokens means extra inference steps and nearly 1.4 times the cost of self-attention. Such additional costs only bring marginal improvements as can be seen in the table.

We also note that TreeTok achieves the shortest average token length among all the tokenizers, which is desirable as \newcite{galle-2019-investigating} shows that \textit{given a fixed vocabulary size budget, the fewer tokens a tokenizer needs to cover the test set, the better the translation.}

\paragraph{Machine Translation.}\label{par:mt}
We conduct experiments on machine translation as a complementary. We use the fairseq framework \footnote{\url{https://github.com/facebookresearch/fairseq/blob/main/examples/translation/README.md##wmt14-english-to-german-convolutional}} to train a Transformer on WMT14 English to German from scratch and measure the performance by calculating the BLEU score on the official test split.

We compare the model's performance when the tokenizer is BPE and TreeTok, respectively. We use the same model training setups. For the two tokenizers, the vocabulary size and basic characters are exactly the same. The results from Table \ref{tbl:renyi} show that TreeTok is slightly better than BPE. 
Based on the results, TreeTok can improve alignment with morphology on top of BPE, while the new segmentation does not compromise downstream task performance.

\subsection{Tree Structure Quality}
\label{sec:tree_quality}
Since tree structures play an important role in both vocabulary construction and segmentation, we evaluate the quality of trees induced by various composition models.
\paragraph{Metric.}
We use recall of morphemes~\cite{van-den-bosch-daelemans-1999-memory} in a tree to assess the quality of the tree structures against gold-standard segmentations, which is defined as the percentage of morphemes in the gold segmentation that can be found in the spans of the evaluated tree.
We discard spans that are trivial for a tree (character-level and word-level spans) and report word-level recall (averaged over word samples).

\paragraph{Baselines.}
For baseline composition models, we include Fast-R2D2~\cite{hu-etal-2022-fast}, which is a variant of DIORA~\cite{drozdov-etal-2019-unsupervised-latent}, and an efficient variant of neural PCFG~\cite{yang-etal-2022-dynamic}. We choose PCFG and R2D2 as our baselines because they represent two different classic approaches to modeling composition: PCFGs are based on underlying grammar structures, while R2D2 (a neural inside algorithm) searches for the optimal information compression structure through binary composition of tokens.

We also include four variants of our composition model for an ablation study. In w/o context, we remove the auto-regression loss from our architecture so that each representation only contains information from individual words. In w/o MorphOverriding, we degenerate $\mathbf{s}_{i,j}$ to the default empty embedding regardless of whether span $\mathbf{x}_{i:j}$ hits the external vocabulary or not. In w/o span loss, for our auto-encoding loss, we only count loss from predicting characters instead of every subword span that hits the external vocabulary. 

\paragraph{Results and Discussions.}
\begin{table}[tb!]
\begin{center}
\setlength{\tabcolsep}{2pt}
\resizebox{0.45\textwidth}{!}{
\begin{tabular}{l|c|c|c}
\hline \hline
                       & Morpho  & Compound  & Word Seg. \\ 
                       & EN.  & EN.  & ZH. \\ \hline
Fast R2D2    & 67.69 & 48.96  & --- \\
Neural PCFG            & 39.87  & 58.33     & 74.26    \\
TreeTok                   & \textbf{90.10} & \textbf{86.20}   & --- \\ \hline
~~~w/o context         & 70.00 & 63.02   & ---\\
~~~w/o \textit{MorphOverriding} & 75.99 & 46.35 & 99.24 \\
~~~w/o span weights     & 89.42            & 78.39 & --- \\
~~~w/o span loss       & 86.79            & 73.70  & --- \\ \hline \hline
\end{tabular}
}
\end{center}
\caption{Performance evaluation of our model, baseline models, and ablation studies on morphological segmentation, measured by morpheme recall rate. EN:English, ZH: Chinese.}
\vspace{-10pt}
\label{tbl:abla}
\end{table}

As shown in Table~\ref{tbl:abla}, our model outperforms all the other composition models. Compared with Fast-R2D2, our main differences lie in the training objectives and the MorphOverriding mechanism. 
This result fully validates the effectiveness of these improvements. Our ablation experiments further analyze the contribution of these improvements to performance enhancement.
Specifically, we have the following findings from each ablation. 
\begin{figure}[tb!]
    \centering
    \includegraphics[width=0.4\textwidth]{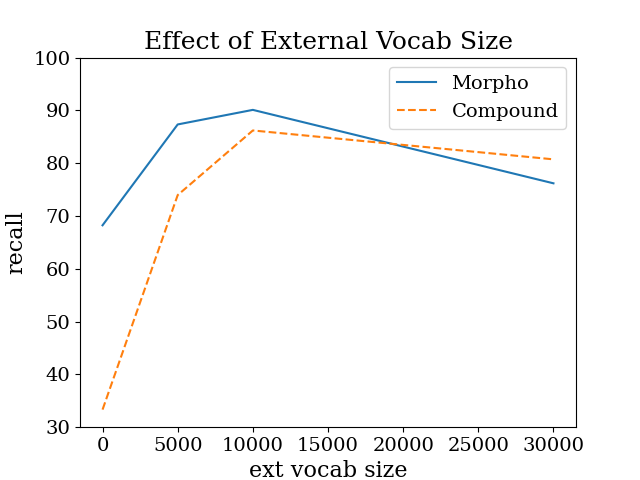}
    \caption{The effect of changing the vocabulary size learned by BPE. The initial results on both tasks show that the performance curve is a concave function where the maximum resides in the middle. }
    \label{fig:vocabsize}
    \vspace{-10pt}
\end{figure}

\begin{table*}[tbh!]
    \centering
    \begin{tabular}{l|c|c|c|c|c} \hline \hline
      \textbf{original word}   & bed  & commonly & windsurfing & tricycles &  uniquenesses \\ \hline
      \textbf{BPE}   & \texttt{bed} & \texttt{commonly} & \texttt{wind/sur/fing} & \texttt{tric/y/cles} & \texttt{uniqu/eness/es} \\
      \textbf{Unigram}   & \texttt{ b/e/d} & \texttt{common/ly} & \texttt{ wind/surf/ing} & \texttt{t/r/i/cycle/s} & \texttt{ unique/ness/e/s} \\
      \textbf{WordPiece} & \texttt{ bed } & \texttt{commonly} & \texttt{ winds/ur/fing} & \texttt{tric/y/cles} & \texttt{ unique/ness/es}  \\
      \textbf{TreeTok} & \texttt{ bed} & \texttt{commonly} & \texttt{ wind/surf/ing} & \texttt{tri/cycles} & \texttt{ unique/ness/es} \\ \hline \hline
    \end{tabular}
    \caption{Example tokenizations.}
    \label{tbl:case_study}
\end{table*}

Removing the auto-regression loss to prevent the model from getting feedback from contextual information significantly impacts the performance on both tasks, especially Morpho. We believe that contextual information can help the model capture the regularities of tenses and learn how to build composition representations for compound words. For example, consider how the context can help determine whether we should build the representation of \texttt{asking} as \texttt{ask+ing} or \texttt{as+king}. While either is a valid combination of morphemes, the former is more likely to be learned by our model since the context around \texttt{asking} often indicates the continuous tense or the gerund form, thus matching better with \texttt{ing}. 

Removing MorphOverriding from the model results in a significant decrease of around 50\% in performance on the decompounding task. The results consolidate our insight about conventional composition models violating the indecomposability of morphemes. Creating a morpheme's representation using its components' representation might make representations of disparate morphemes (e.g., \texttt{wind} and \texttt{win}) entangled together. 

Removing the span loss also causes a performance drop on the two morphology tasks. This aligns well with the insight behind our design of morpheme-level loss, which augments the character-level loss by enhancing the learning of intra-word representations for most morphemes that are at an intermediate granularity.

In addition, we train both Neural-PCFG and our composition model on Chinese wiki~\cite{Xu2019WeaklySD} and evaluate the recall of word boundary against Penn Chinese Treebank~\cite{XUE_XIA_CHIOU_PALMER_2005}. Our model can achieve a word boundary recall of 99.24\% without \textit{MorphOverriding}. Chinese is a language system that can be considered simplistic in terms of its internal structure of words where most of the time, each Chinese character (referred to as a \textit{hàn zì}) represents one morpheme, and there are always explicit boundaries between morphemes without orthographic changes during word formation from characters. These features make compositionality applicable in most cases, thus alleviating the difficulty of modelling intra-word structures. Hence, comparing with the poor performance of w/o \textit{MorphOverriding} on the English dataset (Compound), we can conclude that the difficulties of modelling the internal word structure vary greatly across languages, and \textit{MorphOverriding} is effective and necessary for languages with more challenging morphology structures.

\paragraph{Influence of Heuristic Vocabulary Size} Additionally, we conduct experiments to investigate how the size of our heuristic morpheme vocabulary influences the performance of structure induction.

Figure~\ref{fig:vocabsize} shows that the optimal size of an external vocabulary should be neither too large nor too small. 
According to our hypothesis that the compositional representation of subcomponents of a morpheme should be overridden by a high-level representation, ideally, the external vocabulary should contain all morphemes and only morphemes, because our model will trigger the soft morpheme overriding mechanism for every span that hits the external vocabulary. If BPE is used and the vocabulary is too small, many morphemes (especially longer standalone words) are excluded. Conversely, if it is too large, BPE merges across morphemes, creating spans larger than the smallest meaning-bearing units.

\subsection{Case Studies}
\begin{figure}[tb!]
    \centering
    \includegraphics[width=0.48\textwidth]{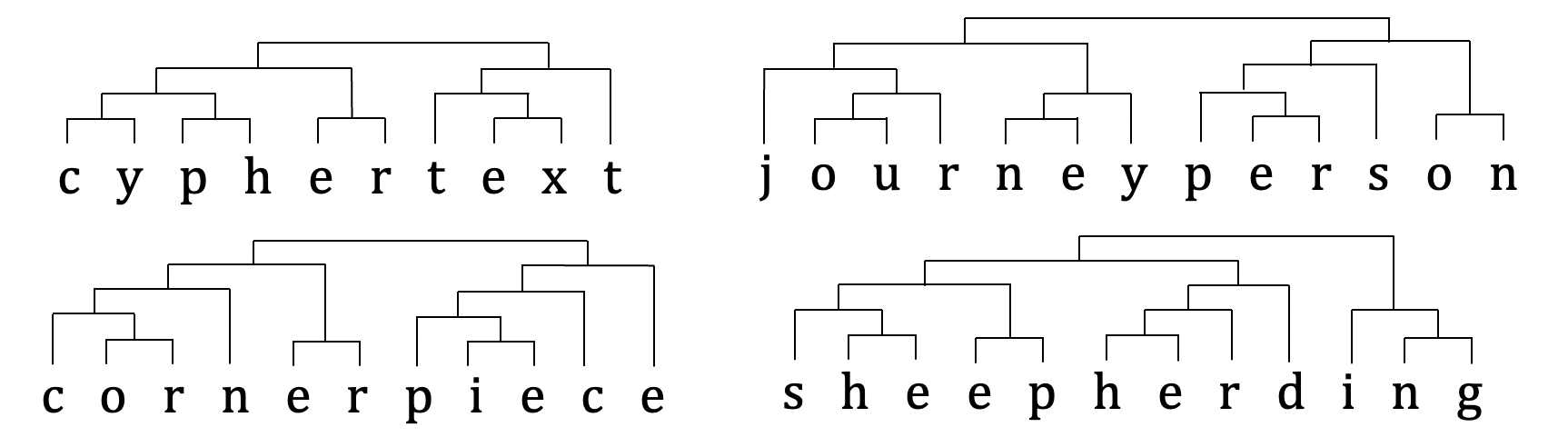}
    \vspace{-10pt}
    \caption{Example tree structures induced by our composition model. }
    \label{fig:tree}
    \vspace{-5pt}
\end{figure}

To further examine the difference between tokenizers, we list their tokenizations in Table~\ref{tbl:case_study} and tree structures induced by our composition model in Figure~\ref{fig:tree}.

Tokens produced by Unigram often include many characters. BPE and WordPiece often violate morpheme boundaries and tokenize words into some intermediate ``junk" tokens introduced during the bottom-up vocabulary construction, such as \texttt{fing}, \texttt{cles}, and \texttt{eness} in Table~\ref{tbl:case_study}.

TreeTok aligns significantly better with morphology. By merging the best of BPE and Unigram pruning, our vocabulary construction algorithm eliminates “junk” tokens. Meanwhile, top-down matching under linguistic constraints prevents excessive word fragmentation and morpheme boundary breaks.

In Figure 4, our model’s high-level tree structures are generally accurate, although some low-level structures appear random, since MorphOverriding prioritizes the most reasonable high-level segmentations based on context, making lower-level details less important.
\section{Conclusion}
Our work introduces the first effective unsupervised neural model for character-level structure induction. We discovered that recognizing the indecomposability of morphemes is key, and to address this, we developed a composition model with a MorphOverriding mechanism alongside two self-supervised objectives. TreeTok induces tree structures that closely match human-labeled morphology and consistently outperforms baselines like BPE and WordPiece across various tasks, offering new insights into unsupervised morphological segmentation.
\section{Limitations}
\begin{table}[htb!]
  \centering
  \resizebox{\columnwidth}{!}{%
    \begin{tabular}{lc}
      \toprule
      Tokenizer & Avg Time/Token (s) \\
      \midrule
      BPE                  & 2.49e-05 \\
      WordPiece            & 2.33e-05 \\
      Unigram              & 2.54e-05 \\
      TreeTok (single processing)     & 1.98e-03 \\
      \bottomrule
    \end{tabular}%
  }%
  \caption{Average processing time per token with identical vocabulary size. TreeTok is run on CPU and tokenises one sample at a time.}
  \label{tab:tokenizer_avg_time}
\end{table}
Our main limitation is that we need additional training and inference overheads. Considering that the composition model only needs to be trained once and the overall time consumption is acceptable\footnote{less than 1 day for 8$\times$ A100 for WikiText-103}, we believe it is not a fatal flaw. Regarding inference cost, because a lightweight parser is produced as a byproduct, it can be afforded by even CPU environments. According to Table \ref{tab:tokenizer_avg_time}, Treetok's average processing time per token is longer than other tokenizers. However, if we allow Treetok to tokenize in batches on a GPU in advance, this gap can be easily compensated. Furthermore, we can maintain a cache of high-frequency words to avoid repeated tokenization. In wikitext-103, the hit rate for a cache that stores the top 100000 frequent words is 98.11\%, which means only 2\% tokens need to be parsed on the fly. e.g. for 1000 tokens, it only needs 0.16s to parse. Furthermore, these results are based on single-core computation, and there is still room for multi-core acceleration.

\section{Acknowledgements}
QZ acknowledges support through the NSF under award 1922658.

\bibliography{anthology,custom}
\bibliographystyle{acl_natbib}

\clearpage
\appendix
\section{Appendix}
\subsection{Pseudo-codes of tokenization}\label{appdx:tokenize}

\begin{algorithm}
\caption{Tokenize}
\small
\begin{algorithmic}[1]
\State Input: string $\mathbf{x}$, parse tree root $r$, vocabulary $\mathbb{V}$
\Procedure{tokenize}{$\mathbf{x}$, $r$, $\mathbb{V}$}
    \State $t \gets []$ \Comment{tokenized subword units list}
    \State $stack \gets [r]$
    \While{$|stack| > 0$}
        \State $c \gets \Call{Pop}{stack}$
        \State $i, j \gets c.i, c.j$
        \State $\bar{\mathbf{x}} \gets \mathbf{x}_{i:j}$ 
        \If{$\bar{\mathbf{x}} \in \mathbb{V}$}
            \State $\Call{Append}{t, \bar{\mathbf{x}}}$ 
        \ElsIf{$i < j$} \Comment{Non-terminal nodes}
            \State $\Call{Push}{stack, c.right}$
            \State $\Call{Push}{stack, c.left}$
        \EndIf
    \EndWhile
    \State $t \gets \Call{PostMerge}{t, \mathbb{V}}$ \\\Comment{post processing if over-split}
    \State \Return $t$
\EndProcedure
\end{algorithmic}
\end{algorithm}

\begin{algorithm}
\small
\caption{Post-Merge Algorithm}
\begin{algorithmic}[1]
\State Input: tokens $\textbf{t}$, vocab2entropy $\mathbb{V}$
\Procedure{PostMerge}{$\textbf{t}$, $\mathbb{V}$}
    \State $n \gets \text{length of } \textbf{t}$
    \If{$n \leq 1$}
        \State $\textbf{t}_\textsc{merge} \gets \textbf{t}$
    \Else
        
        \State $\mathcal{H}[n][n] \text{ init with } \infty $ \Comment{Best entropy}
        \State $s[n][n] \text{ init with } []$ \Comment{Best segments}
        \For{$i \gets 0$ to $n-1$} \Comment{Base case}
            \State $\mathcal{H}_{i,i} \gets \mathbb{V}[x_i]$
            \State $s_{i,i} \gets [x_i]$
        \EndFor
        \For{$h \gets 1$ to $n-1$} \Comment{Iterate tree height}
            \For{$i \gets 0$ to $n - h - 1$}
                \State $j \gets i + h$
                \State $k_\textsc{best} \gets -1$
                \State $m \gets \text{concatenate } t_i\ldots t_j$
                \State $\mathcal{H}_\textsc{best} \gets \Call{Get}{\mathbb{V}, m, \infty}$
                \For{$k \gets i$ to $j-1$}
                    \If{$\mathcal{H}_{i,k}+ \mathcal{H}_{k+1,j} \leq \mathcal{H}_\textsc{best}$}
                        \State $k_\textsc{best} \gets k$
                        \State $\mathcal{H}_\textsc{best} \gets \mathcal{H}_{i,k} + \mathcal{H}_{k+1, j}$
                    \EndIf
                \EndFor
                \If{$k_\textsc{best} \neq -1$}
                    \State $s_{i,j} \gets s_{i,k_\textsc{best}} + s_{k_\textsc{best}+1, j}$
                \Else
                    \State $s_{i,j}\gets [m]$ \Comment{Merge}
                \EndIf
                \State $\mathcal{H}_{i,j} \gets \mathcal{H}_\textsc{best}$
            \EndFor
        \EndFor
        \State $\textbf{t}_\textsc{merge} \gets s_{0,n-1}$
    \EndIf
    \State \Return $\textbf{t}_\textsc{merge}$
\EndProcedure
\end{algorithmic}
\end{algorithm}

\subsection{Pseudo-codes of vocab construction}\label{appdx:vocab}
Please refer to Algorithm \ref{alg:vocab_pruning} for details.
\begin{algorithm}
\caption{Vocabulary Construction}\label{alg:vocab_pruning}
\small
\begin{algorithmic}[1]
\State Input: tree-freq pair list $T$, vocab size $k$, pruning rate $\alpha$
\Procedure{Vocabulary Construction}{$T, k, \alpha$}
    \Procedure{E-step}{$T, \mathbb{V}$}
        \State $\mathbb{V}^{'} \gets \Call{dict}{\ }$ \Comment{E-step: Update vocab freq}
        \For{$\{root, freq\} \in T$}
            \State $\_, seg \gets \Call{TreeViterbi}{root, \mathbb{V}, null}$
            \For{$token \in seg$}
                \State $\mathbb{V^{'}}[token] \gets \mathbb{V^{'}}[token] + freq$
            \EndFor
        \EndFor
        \State \Return $\mathbb{V^{'}}$
    \EndProcedure
    \\
    \Procedure{M-step}{$T, \mathbb{V}$}
        \State $l \gets \Call{dict}{\ }$ \Comment{M-step: Update delta loss}
        \For{$\{root, freq\} \in T$}
            \State $l_{word} \gets \Call{dict}{\ }$ \Comment{word-level delta-loss}
            \State $\_, seg \gets \Call{TreeViterbi}{root, \mathbb{V}, l_{word}}$
            \For{$token \in seg$}
                \State $loss \gets l_{word}[token]$
                \State $l[token] \gets l[token] + loss*freq$
            \EndFor
        \EndFor
        \State \Return $\mathbb{V^{'}}$
    \EndProcedure
    \\
    \State $\mathbb{V} \gets \Call{InitVocab}{T}$ \Comment{Init with a BIG vocab}
    \While{$|\mathbb{V}| > k$}
        \State $\mathbb{V} \gets \Call{E-step}{T, \mathbb{V}}$ \Comment{Estimate token count}
        \State $\mathbb{L} \gets \Call{M-step}{T, \mathbb{V}}$ \Comment{Maximize delta losses}
        \State Remove $\min (|\mathbb{V}|-k, \lfloor \alpha|\mathbb{V}| \rfloor )$ of the 
        \State tokens $t$ with lowest $L_t$ from $\mathbb{V}$
    \EndWhile
    \State \Return $\mathbb{V}$
\EndProcedure
\end{algorithmic}
\end{algorithm}

\begin{algorithm}
\caption{TreeViterbi}\label{alg:viterbi}
\small
\begin{algorithmic}[1]
\State Input: parse tree root $r$, vocabulary $\mathbb{V}$, delta loss dict $l$
\Procedure{TreeViterbi}{$r, \mathbb{V}, l$}
    \State $w \gets r.token$
    \If{$r.i=r.j$}
        \State $s \gets \Call{Get}{\mathbb{V}, w, \infty}$ \Comment{Infinity entropy if $w \notin \mathbb{V}$}
        \State \Return $s, [w]$
    \Else
        \State $s_L, w_L \gets \Call{TreeViterbi}{r.left, \mathbb{V}, l}$
        \State $s_R, w_R \gets \Call{TreeViterbi}{r.right, \mathbb{V}, l}$
        \State $s \gets \Call{Get}{\mathbb{V}, w, \infty}$
        \If{$l$} \Comment{Enter in M-step}
            \State $l[w] \gets l[w] + \Call{Max}{s_L + s_R - s, 0}$ \\\Comment{Record delta loss: Entropy increase}
        \EndIf
        \If{$s_L + s_R > s$}
            \State \Return $s, [w]$
        \Else
            \State \Return $s_L + s_R, w_L + w_R$
        \EndIf
    \EndIf
\EndProcedure
\end{algorithmic}
\end{algorithm}

\begin{algorithm}
\caption{Vocabulary Initialization}\label{alg:vocab_init}
\small
\begin{algorithmic}[1]
\State Input: tree-freq pair list $T$, threshold $k$
\Procedure{InitVocab}{$T, k$}
    \State $\mathbb{V} \gets$ All character freq
    \State $n \gets |\mathbb{V}|$
    \While{True}
        \State $\mathbb{V^{'}} \gets \Call{CountBigrams}{T, \mathbb{V}}$
        \State Prune all the entries in $\mathbb{V^{'}}$ with freq less than $k$
        \State $\mathbb{V}.\Call{merge}{\mathbb{V^{'}}}$ \Comment{Add new items in $\mathbb{V}^{'}$ to $\mathbb{V}$}
        \If{$|\mathbb{V}| = n$}
            \State $break$
        \EndIf
        \State $n = |\mathbb{V}|$
    \EndWhile
    \State \Return $\mathbb{V}$
\EndProcedure
\end{algorithmic}
\end{algorithm}

\begin{algorithm}
\caption{Count Bigrams}\label{alg:count_bigrams}
\small
\begin{algorithmic}[1]
\State Input: tree-freq pair list $T$, vocabulary $\mathbb{V}$
\Procedure{CountBigrams}{$T, \mathbb{V}$}
\State $\mathbb{V^{'}} \gets \Call{dict}{\ }$ \Comment{Store new merges}
    \Procedure{RecurCount}{$r, f$}
        \If {$r.left \And r.right$}
            \State $hit_L \gets \Call{RecurCount}{r.left, f}$
            \State $hit_R \gets \Call{RecurCount}{r.right, f}$
            \If{$hit_L$ and $hit_R$}
                \If{$r.token \in \mathbb{V}$}
                    \State \Return True
                \Else
                    \State $\mathbb{V^{'}}[r.token] \gets f$ \Comment{Merge: new entry}
                    \State \Return False
                \EndIf
            \Else
                \State \Return False
            \EndIf
        \Else
            \State \Return True
        \EndIf

    \EndProcedure

    \For{$\{root, freq\} \in T$}
        \State \Call{RecurCount}{$root, freq$}
    \EndFor
\State \Return $\mathbb{V^{'}}$
\EndProcedure
\end{algorithmic}
\end{algorithm}

\newpage
\subsection{The neural outside pass}\label{appdx:neural_outside}
The outside computation is akin to the inside pass but in a top-down manner. We denote the outside representation and score of a given span as $\bar{\mathbf{o}}_{i,j}^k$ and $\bar{b}_{i,j}^k$ respectively, whose parent span is (i, k) or (k, j) for $k > j$ or $k < i$. 
\begin{equation*}
\scalebox{0.9}{$
\begin{aligned}
&\bar{\mathbf{o}}_{i,j}^k=\left\{\begin{matrix} 
f_{\beta}(\mathbf{o}_{i,k}, \mathbf{i}_{j+1,k}) & \text{ if }k>j \\
f_{\beta}(\mathbf{o}_{k,j}, \mathbf{i}_{k,i-1}) & \text{ if }k<i
\end{matrix}\right. \,,\,\\
&\bar{b}_{i,j}^k=\left\{\begin{matrix}
\phi_{\beta}(\mathbf{o}_{i,k}, \mathbf{i}_{j+1,k})& \text{ if }k>j \\
\phi_{\beta}(\mathbf{o}_{k,j}, \mathbf{i}_{k,i-1})& \text{ if }k<i
\end{matrix}\right. \,,\\
&\check{w}_{i,j}^k=\frac{\exp(\bar{b}_{i,j}^k)}{\sum_{k'>j,k'<i}^{}\exp(\bar{b}_{i,j}^{k'})}\,,\mathbf{o}_{i,j}=\sum_{k > j,k<i}^{}\check{w}_{i,j}^k\bar{\mathbf{o}}_{i,j}^k\,.
\end{aligned}
$}
\end{equation*}
\subsection{Span weights}\label{appdx:span_weights}
An intuitive idea is that the larger the probability of a span's existence, the greater its weight. A span exists if its parent span exists and the span is an immediate child of its parent span. 
Therefore, we can recursively estimate the existence probability of each span top-down~\cite{hu2023a} and formalize the auto-encoding loss as follows:
\begin{equation*}
\scalebox{0.9}{$
\begin{aligned}
&p_{i,j}=\sum_{k<i}^{}p_{k,j} \hat{w}_{k,j}^i + \sum_{k>j}^{}p_{i,k} \hat{w}_{i,k}^j\,,\,p_{1,n}=1\,,\\
&\mathcal{L}_{ae} = -\frac{1}{\sum_{}^{}p_{i,j}}\sum_{\mathbf{x}_{i:j}\in\mathbb{V}}^{}p_{i,j}\log\frac{\exp(\mathbf{o}^T_{i,j}\mathbf{E}_{\mathbb{V}[\mathbf{x}_{i:j}]})}{\sum_{k=1}^{\left | \mathbb{V} \right | }{\exp(\mathbf{o}^T_{i,j}\mathbf{E}_{k})}}\,.
\end{aligned}
$}
\end{equation*}

\subsection{Experimental Setup and Hyperparameters}\label{appdx:model_config}
Our composition function uses 4 layers of Transformer layers. For span representations, we use 128-dimensional embeddings with 4 attention heads, 512-dimensional hidden layer representations, and a vocabulary size of 7835. This vocabulary is built from concatenating 1903 most frequent characters in the training set of wikitext-103 and a 10,000-entry BPE dictionary, excluding all characters. To guide the composition function, our lightweight parser is a 4-layer Transformer model that uses 64-dimensional embeddings with 4 attention heads and 128-dimensional hidden layer representations. For the causal language model, we use a 3-layer GPT2 equipped with 128-dimensional embeddings and 4 attention heads and follow the original configuration for the rest of the hyperparameters.

Our composition models are trained on 8 PPUs with a learning rate of 1e-2 for the light-weight parser and 5e-4 for the rest.
The batch size is $8\times128$, and for each sample, we limit the context window to 512 characters (whitespace included). The total number of training steps is ten times the number of sentences in Wikitext-103.

\end{document}